\documentclass[conference]{IEEEtran}
\ifCLASSINFOpdf
  \usepackage[pdftex]{graphicx}
\else
 \usepackage[dvips]{graphicx}
\fi
\usepackage{algorithm}
\usepackage{algorithmic }
\usepackage{float}
\usepackage{array}

\begin{document}
\title{Learning to Shoot in First Person Shooter Games by Stabilizing Actions and Clustering Rewards for Reinforcement Learning}

\author{\IEEEauthorblockN{Frank G. Glavin}
\IEEEauthorblockA{College of Engineering \& Informatics,\\
National University of Ireland, Galway,\\
Ireland.\\
Email: frank.glavin@nuigalway.ie}
\and
\IEEEauthorblockN{Michael G. Madden}
\IEEEauthorblockA{College of Engineering \& Informatics,\\
National University of Ireland, Galway,\\
Ireland.\\
Email: michael.madden@nuigalway.ie}
}


%Telephone: (800) 555--1212, Fax: (888) 555--1212}


\maketitle

\begin{abstract} 
While reinforcement learning (RL) has been applied to turn-based board games for many years, more complex games involving decision-making in real-time are beginning to receive more attention. A challenge in such environments is that the time that elapses between deciding to take an action and receiving a reward based on its outcome can be longer than the interval between successive decisions. We explore this in the context of a non-player character (NPC) in a modern first-person shooter game. Such games take place in 3D environments where players, both human and computer-controlled, compete by engaging in combat and completing task objectives. We investigate the use of RL to enable NPCs to gather experience from game-play and improve their shooting skill over time from a reward signal based on the damage caused to opponents. We propose a new method for RL updates and reward calculations, in which the updates are carried out periodically, after each shooting encounter has ended, and a new weighted-reward mechanism is used which increases the reward applied to actions that lead to damaging the opponent in successive hits in what we term ``hit clusters''.
\end{abstract} 


% \fi
\IEEEpeerreviewmaketitle

\section{Introduction}
In this work, we consider situations in which artificial intelligence (AI) agents acquire the necessary skills to play a game through trial and error. Reinforcement learning (described briefly below) is a suitable paradigm to enable an AI agent to learn in parallel with their opponents from in-game experience and has long been applied to games, particularly turn-based board games.

Modern first-person shooter (FPS) games (also described briefly below) operate in real-time, take place in three-dimensional environments that are detailed and complex, and involve humans playing with or against computer-controlled NPCs, which introduces interesting challenges for RL. For example, an RL agent may need to make a rapid sequence of action choices, and the time taken for an action to result in a reward is longer than the time interval between successive actions. Thus, a naive application of a standard RL approach would result in rewards not being allocated to the appropriate originating actions.

\subsection{Reinforcement Learning}
The basic underlying principle of reinforcement learning \cite{sutbar} is that an \emph{agent} interacts with an \emph{environment} and receives feedback in the form of positive or negative rewards based on the actions it decides to take. The goal of the agent is to maximize the long term reward that it receives. For this, a set of states, actions and a reward source must be defined. The \emph{state space} comprises a (typically finite) set of states, representing the agent's view of the world. The \emph{action space} comprises all of the actions that the agent can carry out when in a given state. The \emph{reward signal} provides the agent with either a reward or a penalty (negative reward) depending on how successful the action was and which was carried out in the state. State-action pairs are recorded which represent the expected value of carrying out an action in a given state and these represent the \emph{policy} of the agent. 

This research involves incorporating reinforcement learning into the logic of an FPS bot to enable it to learn the task of shooting. We will now take a look at the FPS genre of computer games, the role of non-player characters and the motivations for this work.

\subsection{First Person Shooter Games and Non-Player Characters}
First person shooter games take place in a 3D world in which the player must battle against opponents, from a first-person perspective, and complete game objectives. The game environment includes ``pickups'' such as weapons, ammunition and health packages. Human players must learn the pros and cons of using each weapon, familiarize themselves with the map and master the game controls for navigation and engaging in combat. There are several different game types to choose from with the most basic being a \emph{Deathmatch} where each player is only concerned with eliminating all other players in the environment. Players are spawned (their avatar appears) on the map with basic weaponry and must gather supplies and engage in combat with other players. The game finishes when the time limit has elapsed or the score limit has been reached. Objective-based games also exist such as \emph{Domination} in which players have to control specific areas of the map in order to build up points. \\
\indent NPCs are computer-controlled players that take part in the game and traditionally have scripted rules to drive their behavior. They can be programmed with certain limitations in order to produce a game experience that has a specified difficulty level for a human player. These limitations can include timed reaction delays and random perturbation to its aim while shooting.
\subsection{Motivation}
We believe that FPS games provide an ideal testbed for carrying out experimentation using reinforcement learning. The actions that the players carry out in the environment have an immediate and direct impact on the success of their game play. Decision-making is constant and instantaneous with players needing to adapt their behavior over time in an effort to improve their performances and outwit their opponents. Human players can often spot repetitive patterns of behavior in NPCs and adjust their game-play accordingly. This predictive behavior of traditional NPCs can result in human players losing interest in the game when it no longer poses a challenge to them. Lack of adaption can often result in the NPC being either too difficult or too easy to play against depending on the ability of the human player. We hypothesize that enabling the bot to learn how to play competently, based on the opponents movements, and adapt over time will lead to greater variation in game-play, less predictable NPCs and ultimately more entertaining opposition for human players. \\
\indent For the research presented here, we are concentrated solely on learning the action of shooting a weapon. We believe that shooting behavior should be continually adaptive and improved over time with experience in the same way a human player would learn how to play.

\section{Related Research}
General background information on the use of AI in virtual agents can be found in Yannakakis and Hallam \cite{yann2007}. A wide variety of computational intelligence techniques have been successfully applied to FPS games such as Case-Base Reasoning (CBR) \cite{auslander}, Decision Trees \cite{leiva2011}, Genetic Algorithms \cite{esparcia} and Neural Networks \cite{petrakis}. In this section we will discuss our previous research and contrast our shooting approach to that of others form the literature. \\
\indent Our research is concerned with applying reinforcement learning to the behavior of virtual agents in modern computer games. In our earlier work, we developed an FPS bot, called DRE-Bot \cite{fg1}, which switches between three high-level modes of \emph{Danger}, \emph{Replenish} and \emph{Explore}, each of which has their own individual reinforcement learner for choosing actions. We designed states, actions and rewards specifically for each mode with the modes being activated based on the circumstances of the game for the bot. The inbuilt shooting mechanism from the game was used in this implementation and we carried out experimentation against scripted fixed-strategy bots. Our findings showed that the use of reinforcement learning produced varied and adaptable NPCs in the game. We later developed the RL-Shooter bot \cite{fg2} which uses reinforcement learning for adapting shooting over time based on a dynamic reward from the opponent damage values. We carried out experimentation against different levels of fixed-strategy opponents and discovered a large amount of variance in the performances, however, there was not a clear pattern of the RL-Shooter bot continuing to improve over time. These findings led to our current research in which our aim is to show clear evidence of learning in shooting performance over time.\\
\indent The use of AI methodologies to control NPCs in FPS games has received notable attention in recent years with the creation of the \emph{Bot Prize} \cite{bot1} competition. This competition was set up in 2008 for testing the humanness of computer-controlled bots in FPS games. Two teams, \emph{MirrorBot} and \emph{UT\textsuperscript{2}} (described below), surpassed the humanness barrier of 50 percent in 2012. As mentioned earlier, we are concentrating on one specific NPC game task at the moment which will eventually form part of a general purpose bot that we would hope to submit to such a competition in the future.\\
\indent MirrorBot \cite{mirror2013} functions by recording opponents' movements in real-time. If it detects what it perceives to be a non-violent player it will proceed to mimic the opponent by playing back the recorded actions, with slight differences, after a short delay. The purpose of this is to give the impression that the bot is independently selecting the actions. In our shooting implementation, the bot will actually be deciding the shooting actions to take based on what has worked the best from its experience. MirrorBot shoots by adjusting it's orientation to a given focus location. It also anticipates the opponents movements by shooting at a future location based on the opponents velocity. The authors do not report that this technique is improved or adapted over time and therefore may become predictable to experienced players. The weapon selection decision for MirrorBot is based on the efficiency of the weapon and the amount of currently available ammunition. \\ 
\indent The UT\textsuperscript{2} bot \cite{schrum} uses data collected from human traces of navigation when it detects that its own navigation system has failed. The bot also uses a combat controller with decision-making that was evolved using artificial neural networks. The bot shoots at the location of the opponent with some random added noise based on the relative velocity and distance from them. This again differs from our shooting architecture as we are emulating the process of a human player learning and becoming more proficient with experience. We believe that such a characteristic is essential to create truly adaptive NPC opponents.\\
\indent Van Hoorn \emph{et al.} \cite{van2009} developed a hierarchical learning-based architecture for controlling NPCs in Unreal Tournament 2004. Three sub-controllers were developed for the tasks of combat, exploration and path-following and these were implemented as recurrent neural networks that were trained using artificial evolution. Two fitness functions were used when evolving the shooting controller. One of these measured the amount of damage the agent caused and the other measured the hits-to-shots fired ratio. Once the sub-controllers have been evolved to a suitable performance level their decision-making is ``frozen'' and they no longer evolve during the game-play. This contrasts with our approach which is based on consistently adaptive in-game behavior based on real-time feedback from the decision-making. \\
\indent McPartland and Gallagher \cite{ref7} created a purpose-built FPS game and incorporated the tabular SARSA($\lambda$) \cite{sutbar} reinforcement learning algorithm into the logic of the NPCs. Controllers for navigation, item collection and combat were individually learned. Experimentation was carried out involving three different variations of the algorithm, namely, \emph{HierarchicalRL}, \emph{RuleBasedRL} and \emph{RL}. HierarchicalRL uses the reward signal to learn when to use each of the controllers. The RuleBasedRL uses the navigation and combat controllers but has predefined rules on when to use each. The RL setup learns the entire task of navigation and combat together by itself. The results showed that reinforcement learning could be successfully applied to the simplified purpose-built FPS game. Our shooting implementation also uses the SARSA($\lambda$) algorithm, described later in Section \ref{sars}, to drive the learning of the NPC, however the two architectures are very different. Firstly, we are deploying an NPC into a commercial game over a client-server connection as opposed to a basic purpose-built game. We are also only concerned with learning the task of shooting by designing states from the first-person perspective of the NPC and reading feedback from the system based on damage caused to opponents. \\
\indent Wang and Tan \cite{falc} used a self-organizing neural network that performs reinforcement learning, called \emph{FALCON}, to control NPCs in Unreal Tournament 2004. Two reinforcement learning networks were employed to learn both behavior modeling and weapon selection. Experimentation that was carried out showed that the bot could learn weapon-selection to the same standard as hard-coded expert human knowledge. The bot was also shown to be able to adapt to new opponents on new maps if its previously learned knowledge was retained. The implementation used the inbuilt shooting command with random deviations added, to the direction of the shooting, in an effort to appear human-like. This architecture, while using reinforcement learning for other aspects, does not learn how to improve shooting over time and will just randomly deviate the aim from the opponent.
\section{Methodology}
\subsection{Development Tools}
We developed the shooting architecture for the game Unreal Tournament 2004 (UT2004) using an open-source development toolkit called Pogamut 3 \cite{ref6}. UT2004 is a commercial first person shooter game that was developed primarily by Epic Games and Digital Extremes and released in 2004. It is a multi-player game that allows players to compete with other human players and/or computer-controlled bots. UT2004 is a highly customisable game with a large number of professionally-made and user-made maps, a wide variety of weaponry and a series of different game-types from solo play to cooperative team-based play. The game was released over ten years ago and, although computer game graphics in general have continued to improve since then, the FPS formula and foundations of game-play remain the same in current state-of-the-art FPS games. Pogamut 3 makes use of \emph{UnrealScript} for developing external control mechanisms for the game. The main objective of Pogamut 3 is to simplify the coding of actions taken in the environment, such as path finding, by providing a modular development platform. It integrates five main components: Unreal Tournament 2004, GameBots2004, the GaviaLib Library, the Pogamut Agent and the NetBeans IDE. A detailed explanation of the toolkit can be found in Gemrot \emph{et al.} \cite{ref6}.
\subsection{RL Shooting Architecture Details}
We designed the RL shooting architecture to enable the bot to learn how to competently shoot a single weapon in the game. The weapon chosen was the \emph{Assault Rifle}. This weapon, with which each player is equipped when they spawn, is a machine gun that is most effective on enemies that are not wearing armour, and it provides low to moderate damage. The secondary mode of the weapon is a grenade launcher. We use a ``\emph{mutator}'' to ensure that the only weapon available to the players on the map is the Assault Rifle; this is a script that changes all gun and ammunition pickups to that of the Assault Rifle when it is applied to the game. The architecture is only concerned with the primary firing mode of the gun in which a consistent spray of bullets is fired at the target. We chose this design to enable us to closely analyse and view the trend of performance and learned behavior over time. Actions could, of course, be tailored for both modes of each weapon in the game. The game also has a slight inbuilt skew for the bullets to imitate the natural recoil of firing the gun. We hypothesized at the outset that the bot will still be able to learn in the presence of this recoil and will adjust its technique as a human player would. 

\subsubsection{States}
\begin{figure}[h]
  \begin{center}  
      \includegraphics[width=1.5in]{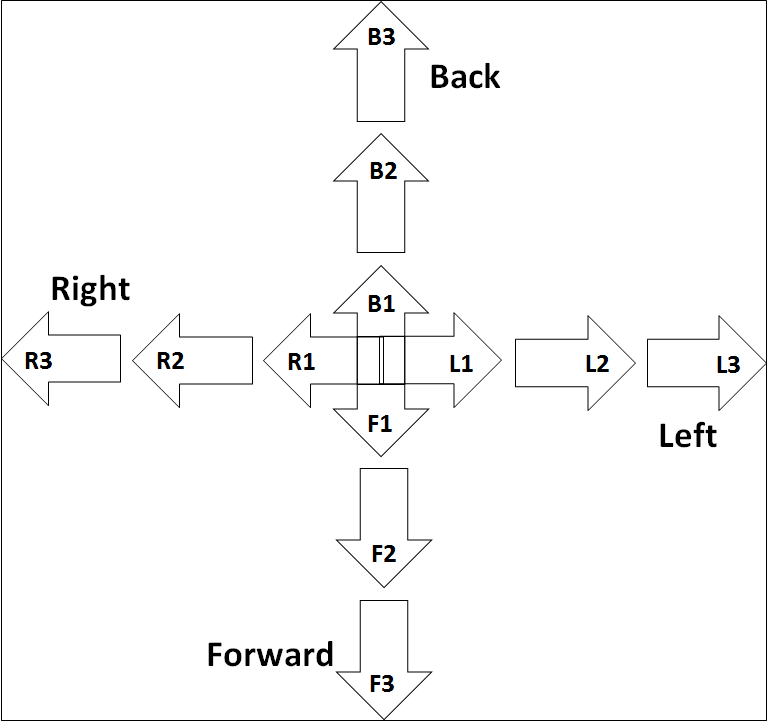}
    \caption{The relative velocity values for the state space.}
    \label{vel}
  \end{center}
\end{figure}

The states are made up of a series of checks based on the relative position and speed of the nearest visible opponent. Specifically, the relative speed, relative moving direction, relative rotation and distance to the opponent are measured. The velocity and direction values of the opponent are read from the system after being translated into the learner bots point of view. The opponent's direction and speed are recorded relative to the bot's own speed and from the perspective of the learner bot looking directly ahead. The opponent can move forwards (F), backwards (B), Right (R) or Left (L) relative to the learner bot, at three different discretized speeds for each direction as shown in Figure \ref{vel}.

\begin{figure}[h]
  \begin{center}  
      \includegraphics[width=2in]{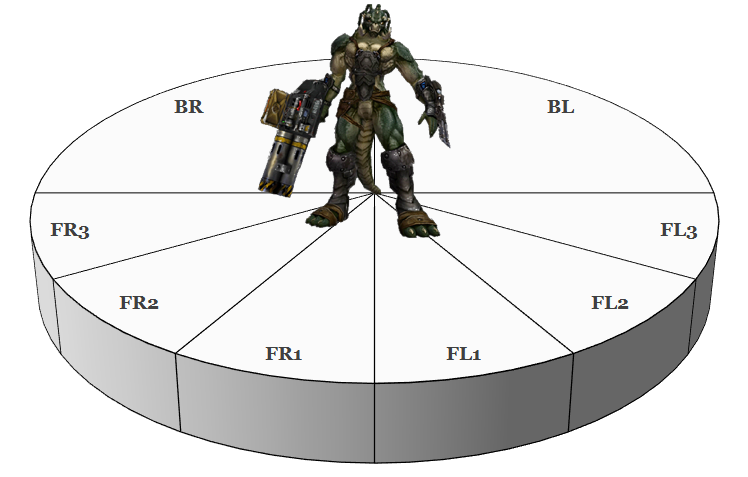}
    \caption{The relative rotation values for the state space.}
    \label{rot}
  \end{center}
\end{figure}

UT 2004 has its own in-game unit of measurement called an \emph{Unreal Unit} (UU) that is used in measuring distance, rotation and velocity. In our velocity state representation, Level 1 is from 0 to 150 UU/sec, Level 2 is from 150 to 300 UU/sec and Level 3 is greater than 300 UU/sec. The bot can be moving in a combination of forward or backward and left or right at any given time. For instance, one state could be R3/F1 in which the opponent is moving quickly to its right while slowly moving forward. There are six forward/backward moving states and six left/right moving states. The bot being stationary is another state so there are thirty seven possible values for the relative velocity states. \\
\indent The direction that the opponent is facing (rotation) relative to the bot looking straight ahead is also recorded for the state space. This is made up of eight discretized values. There are two back facing rotation values which are Back-Left (BL) and Back-Right (BR). There are six forward facing rotation values, three to the left and three to the right, with each consisting of thirty degree segments as shown in Figure \ref{rot}.

The distance of an opponent to the bot is also measured and is discretized into the following values: ``close'', ``regular'', ``medium'' and ``far'' as shown in Table \ref{distValues}. These values are map-specific and were determined after observing log data and noting the distributions of values for opponent distances that were recorded. \\

\begin{table}[h]
\caption{Discretized Distance Values}
\begin{center}
\setlength{\extrarowheight}{0.15cm}
\begin{tabular}{|c|c|}
   \hline
   \bf{State}  & \bf{Distance} \\
   \hline
   \emph{Close} & 0 - 500 UT \\
 \hline
\emph{Regular} & 500 - 1000 UT \\
 \hline
   \emph{Medium} & 1000 - 1500 UT \\
\hline
   \emph{Far} & \textgreater 1500 UT \\
\hline
 \end{tabular}
\label{distValues}
\end{center}
\end{table}

\indent The state space was specifically designed to provide the bot with an abstract view of the opponents movements so that an informed decision can be made when choosing what direction to shoot, with the goal that over time the bot will learn the most effective shooting technique based on the circumstances that it finds itself in. This state space representation could also be used to design a learner bot for the other weapons in the game, by developing suitable actions, as it encompasses the general movements of an opponent from the first-person perspective of the bot.

\subsubsection{Actions}
\begin{figure*}[ht!]
\begin{center}
      \includegraphics[width=5.0in]{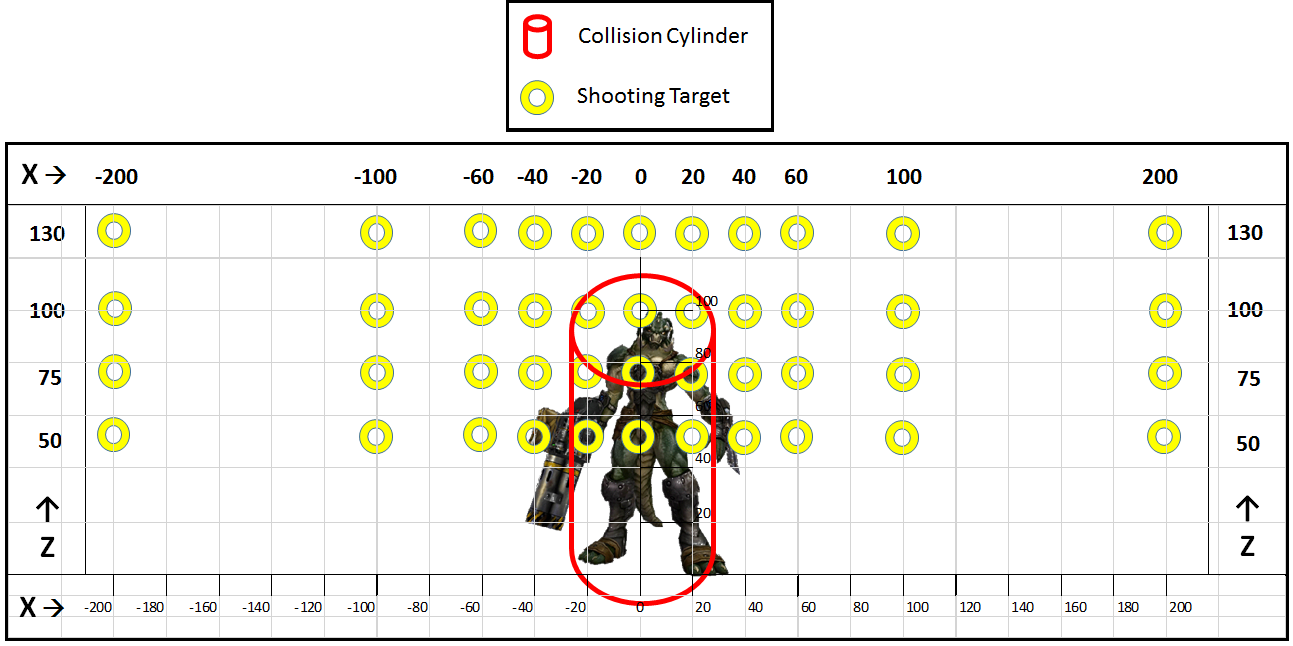}
    \caption{A visualisation of the shooting actions available to the bot.}
    \label{shootingActionsNew}
\end{center}
\end{figure*}

The actions that are available to the bot are expressed as different target directions in which the bot can shoot, and which are skewed from the opponent's absolute location on the map. The character model in the game stands approximately 50 UU wide and just under 100 UU in height and can move at a maximum speed of 440 UU/sec while running. (The system can, however, record higher velocity values on occasion when the bot receives the direct impact of an explosive weapon such as a grenade.) The amount of skew along the X-axis (left and right) and Z-axis (up and down) varies by different fixed amounts as shown in Figure \ref{shootingActionsNew}. The Z-axis skews have four different values which range from the centre of the opponent to just above its head. The X-axis skews span from left to right across the opponent with the largest skews being 200 UU to the left and 200 UU to the right. These actions were designed specifically for the Assault Rifle weapon.
\subsubsection{Rewards}
The bot receives a reward of 250 every time the system records that it has caused damage to the opponent with the shooting action. If the bot shoots the weapon and does not hit the opponent it receives of penalty of -1. These values were chosen as they produced the best results during a series of parameter search runs in which the reward value was modified. The reward is adjusted depending on the proximity of hits to other hits when the PCWR technique (which will be described later) is enabled. If the technique is disabled then the bot will receive 250 for every successful hit.

\subsection{SARSA($\lambda$) Algorithm}
\label{sars}
The shooting architecture uses the SARSA($\lambda$)\cite{sutbar} reinforcement learning algorithm with two variations to its conventional implementation. The bot calls the \emph{logic} method (which drives its decision-making) approximately every quarter of a second. This is required so that the bot is capable of reacting in real-time. It is possible to adjust how often the bot calls the logic method but if it is increased to half a second or a second then the bot becomes visibly less responsive. We have identified that if the bot is selecting four shooting directions a second there is sometimes a credit assignment problem in which the reward for a successful hit will be incorrectly assigned to a different action that was selected after the initial action that caused the damage. This is due to a delay in the time it takes to register a hit on the opponent after selecting an action. We address this problem through the use of Persistent Action Selection, discussed later, by ensuring that when an action is chosen, it remains the chosen action for a set number of time steps before a new action is chosen. The algorithm is still run four times a second with the perceived state changing at the this rate, however, the action-selection mechanism is set up to repeatedly select the same action in groups of three time steps. The states, and their corresponding state-action values, will continue to change at each time step but the action chosen will remain the same over three time steps. The reward can also be adjusted by PCWR as described in the next section. \\
\indent We initialized all of the values in the state-action and eligibility trace tables to zero and used the following parameters for the SARSA($\lambda$) algorithm. The learning rate, $\alpha$, determines how quickly newer information will override older information. We would like the bot to have strong consideration for recent information without completely overriding what has been learned so this value is set to 0.7. The discount parameter, $\gamma$, determines how important future rewards are. The closer the value is to 0, the more the agent will only consider current rewards whereas a value close to 1 would mean the agent would be more focused on long term rewards. To enable a balance between current and longterm rewards we set the value of $\gamma$ to 0.5. The eligibility trace, $\lambda$, is set to 0.9. This value represents the rate at which the eligibility traces decay over time. Setting this as a large value results in recent state-action pairs receiving a large portion of the current reward. The $\epsilon$-greedy action-selection policy is used with the exploration rate initialized at 20\%. This is reduced by 3\% every one hundred deaths and when it reaches 5\% it remains at this level. This percentage determines how often the bot will explore new actions as opposed to exploiting previously learned knowledge. During the early stages of learning we encourage exploration of different shooting strategies. We do not reduce exploration below 5\% to ensure the behavior does not become predictable and that new strategies can be explored a small percentage of the time later in the learning process. A detailed explanation of the SARSA($\lambda$) algorithm can be found in Sutton and Barto \cite{sutbar} and our previous work \cite{fg2}. In our current shooting implementation, the states and actions for each step are stored and the updates are carried out in sequence once the shooting period has ended. This enables us to shape the reward using PCWR if required. The process is illustrated in Figure \ref{PCWR}.
%
%\begin{algorithm}
%\caption{Pseudocode for the Sarsa($\lambda$) algorithm.}
%\begin{algorithmic}
%\FORALL{\emph{s}, \emph{a}}
%\STATE \emph{Q}(\emph{s}, \emph{a}) = 0
%\STATE \emph{e}(\emph{s}, \emph{a}) = 0
%\ENDFOR
%\REPEAT 
%\STATE Initialize \emph{s}, \emph{a}
%\REPEAT
%\STATE Take action \emph{a}, observe \emph{r}, \emph{s'}
%\STATE Choose \emph{a'} and \emph{s'} using policy derived from \emph{Q} 
%\STATE $\delta$ $\Leftarrow$ \emph{r} + $\gamma$\emph{Q}(\emph{s'}, \emph{a'}) - \emph{Q}(\emph{s}, \emph{a})
%\STATE \emph{e}(\emph{s}, \emph{a}) $\Leftarrow$ 1
%\FORALL{\emph{s}, \emph{a}}
%\STATE \emph{Q}(\emph{s}, \emph{a})  $\Leftarrow$ \emph{Q}(\emph{s}, \emph{a}) + $\alpha$$\delta$\emph{e}(\emph{s}, \emph{a})
%\STATE \emph{e}(\emph{s}, \emph{a}) $\Leftarrow$ $\gamma$$\lambda$\emph{e}(\emph{s}, \emph{a})
%\ENDFOR
%\STATE \emph{s} $\Leftarrow$ \emph{s'}; \emph{a} $\Leftarrow$ \emph{a'}
%\UNTIL{(steps of single episode have finished)}
%\UNTIL{(all episodes have finished)} 
%\end{algorithmic}
%\label{sarsaAlg}
%\end{algorithm} 
\subsection{Periodic Cluster-Weighted Rewarding}
\begin{figure*}[ht!]
\begin{center}
  \includegraphics[width=5in]{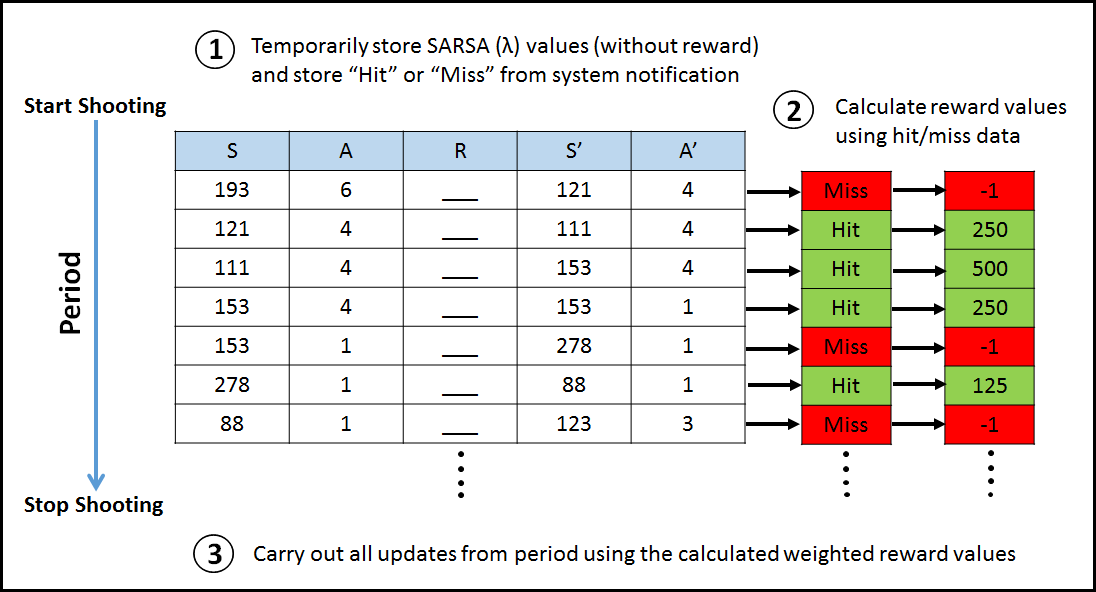}
  \caption{An illustration of how Periodic Cluster-Weighted Rewarding works.}
\label{PCWR}
\end{center}
\end{figure*}

Periodic Cluster-Weighted Rewarding (PCWR) involves weighting the percentage of reward that is applied to an action that successfully caused a damage ``hit'' to an opponent, while hits that occur in clusters of other hits will receive greater reward. If a single standalone hit occurs then the action receives half of the reward. The purpose of this is to promote behavior that is indicative of good FPS game play. This is achieved by providing more reinforcement to actions which resulted in groups of hits on the opponent. If a number of hits occur in a row, the two outermost (the hits that occur next to misses) receive the full 250 reward\footnote{The value 250 was chosen for the full reward as it produced the best results in preliminary runs in which various values were tested between 1 and 1000.}. All of the other hits inside the cluster receive double the reward value (500). Suppose, for example, that we have a sequence of eight actions chosen, and the recordings for each are as follows: 1:Miss, 2:Hit, 3:Hit, 4:Hit, 5:Miss, 6:Hit, 7:Miss. Actions 2 and 4 will receive a reward of 250 having occurred as the outer actions of the cluster with action 3 receiving 500 (double the reward). The standalone action 6 will receive 125 (half of the reward). This process is illustrated in Figure \ref{PCWR}. The purpose of this is to increase the positive reinforcement of actions that lead to the bot causing a significant amount of damage to the opponent. From when the bots starts shooting to when it stops we have termed a shooting ``period''. During this period, all of the states, actions and hit/miss values are recorded as they happen. Once the shooting period ends, all of the Q-table updates are carried out in sequence with the cluster weighting having been applied to the reward value.
\subsection{Persistent Action Selection}
As mentioned earlier, the bot reads information about its current state from the system four times a second. These short time intervals are required to enable the bot to perceive and react to situations in real-time. Persistent Action Selection (PAS) involves choosing an action and then keeping this as the selected action over multiple time steps. The states are still changing and being read every 0.25s but the actions that are selected are being persisted over multiple time steps. The purpose of this is to minimize the occurrences of mis-attribution of reward in a setting where a new action could have been selected before the reward was received for the previous action. Persisting with the same action also naturally amplifies the positive or negative reinforcement associated with that particular action. If it is an action that does not lead to any hits then it will be less likely chosen in the future. Although the actions persist over \emph{n} time steps, the bot's response time will continue to be every 0.25s. Therefore, since actions are specified relative to the current location of the opponent, the shooting direction will change as the opponent moves, even when the action is being persisted. Intervals in the range of 2 to 10 were all tested and the value 3 was chosen as it produced the best hit accuracy and the shooting behavior looked most natural to human observers.
\subsection{Discussion}
This shooting architecture is, of course, only as efficient as the manual design of states, actions and rewards for the specific task will allow. What it does provide, however, is a real-time adaption of shooting based on in-game experience and a knowledge base timeline as the learner progresses (intermittent Q-value tables can be stored offline). The architecture works as a standalone system that could potentially be ``plugged-in'' to other existing bot projects which address the tasks of navigation, item collection etc. It is novel in that it will continually change its shooting technique based purely on the success or failure of past actions in the same circumstances. This will prevent the bot from appearing predictable to human opposition.

\section{Experimentation}
The experimentation that is described in this section includes the analysis of four different variations of the shooting architecture. These include PCWR enabled with PAS enabled over three time steps, PCWR enabled with actions selected at every time step, PCWR disabled with PAS enabled over three time steps and PCWR disabled with actions selected every time step. From here on, these will be abbreviated as \emph{PCWR:Yes\textunderscore PAS:Yes}, \emph{PCWR:Yes\textunderscore PAS:No}, \emph{PCWR:No\textunderscore PAS:Yes} and \emph{PCWR:No\textunderscore PAS:No} respectively. Ten individual runs were carried out for each of these variations in order to analyse the averaged results. All of the games were played out in real-time.
\subsection{Details}
Each run involves the RL bot (one of four variations as described earlier) taking part in a Deathmatch game against a single Level 3 (Experienced) fixed-strategy opponent and continually playing until it has died 1500 times. All of the runs took place on the \emph{Training Day} map. This is one of the default maps from the game that is designed for two to three players and encourages almost constant combat. The map was chosen specifically for this reason, to minimize the amount of time that players would spend searching for each other. The only gun available for the duration of the game is the \emph{Assault Rifle}, which is a low powered weapon that shoots a consistent flow of bullets, and which each player is equipped with when they appear on the map. The accumulation of kills and deaths are recorded and the number of hits, misses and rewards per life are also recorded. 
\subsection{Results and Analysis} 
Table \ref{averageValues} shows the average hits, misses and rewards per life that the bot achieved when PCWR and PAS were both enabled and disabled. The values shown are averaged over 10 runs in each case. The first observation that we can make is that the average hits per life is much greater when the actions are persisted over 3 time steps. When PAS is enabled the bot achieves almost double the number of hits per life whereas there is little change in the number of misses per life.
\begin{table}[t]
\caption{Averages per life for hits, misses and rewards.}
\label{averageValues}
\begin{center}
\setlength{\extrarowheight}{0.15cm}
\begin{tabular}{|c|c|c|c|}
   \hline
\bf{Technique}&\bf{Hits} &\bf{Misses} &\bf{Reward}  \\
\hline
 \emph{PCWR:Yes\textunderscore PAS:Yes} & 25.64 &40.94 &3753.60 \\
 \emph{PCWR:No\textunderscore PAS:Yes} & 26.22 &42.18&6512.72 \\
   \emph{PCWR:Yes\textunderscore PAS:No} & 12.99 &39.31&1530.19 \\
   \emph{PCWR:No\textunderscore PAS:No} &13.39&39.48&3307.29\\
\hline
\end{tabular}
\end{center}
\end{table}
We can also see that the average total reward per life is almost halved when PCWR is enabled. This is the result of a large number of isolated hits occurring during the games in which the bot only receives 50\% of the reward for the hit. Since the reward scheme is different when PCWR is enabled and disabled, reward averages would not be expected to be within a similar range.\\
\indent The overall percentage shooting accuracy for each scenario is listed in Table \ref{accLives}. This accuracy is averaged over 10 runs for each and is calculated as the hits divided by the total shots taken. When PCWR is enabled using PAS, the bot achieves slightly better accuracy than when it is disabled. We can once again see a large difference in performance between the runs where actions are selected every time step and those that persist over 3 time steps. This table also lists the best kill streak achieved by each of the bots over all of the runs. A kill streak is achieved when the bot kills an opponent several consecutive times without dying itself. Although we have not designed the bot to maximize the skill streak that it can achieve, we can observe that it is a direct result of learning to proficiently user the weapon. The PCWR:Yes\textunderscore PAS:Yes bot achieved a maximum kill streak of 15 on 3 of 10 game runs. The PCWR:No\textunderscore PAS:Yes bot achieved a maximum kill streak of 14 whereas the two bots that choose an action every time step both only managed to reach a maximum kill streak of 7. The final information from this table is the average total time alive for each of the bots for the 1500 lives. Proficient shooting skills result in the bot staying alive for longer and there is an average difference of between 4 and 5 hours when actions are persisted over 3 time steps as opposed to being selected every time step. \\
\begin{table}[t]
\caption{Overall average percentage accuracy, maximum kill streak and average hours alive per game.}
\label{accLives}

\begin{center}
\setlength{\extrarowheight}{0.15cm}
\begin{tabular}{|c|c|c|c|}
\hline
\bf{Technique}&\bf{Accuracy}  &\bf{Kill Streak} & \bf{Hours Alive}\\
\hline
\emph{PCWR:Yes\textunderscore PAS:Yes} &38.51\%&15&17.55 \\
\emph{PCWR:No\textunderscore PAS:Yes} & 38.33\% &14&17.95 \\
\emph{PCWR:Yes\textunderscore PAS:No} & 24.84\% &7&12.90\\
\emph{PCWR:No\textunderscore PAS:No} &25.32\%&7&13.04\\
\hline
\end{tabular}
\end{center}
\end{table}
Table \ref{kdRatioStats} shows the average, minimum and maximum final kill-death ratio after 1500 lives over 10 runs. The kill-death ratio is calculated by dividing the number of kills the bot achieves by the number of times it dies. For instance, if the bot kills 5 times and has died 4 times then its kill-death ratio would be 1.25:1. On average the bots that use PAS over 3 time steps kill the opponent over twice as often as they die. The two bots that select a new action every time step always die more times than they are able to kill. The single best kill-death ratio overall was achieved by the PCWR:Yes\textunderscore PAS:Yes bot, however, it performs slightly worse on average than the PCWR:No\textunderscore PAS:Yes bot. Kill-death ratio gives a good high-level indication of performance in FPS games.

\begin{table}[t]
\caption{Average, minimum and maximum final kill-death ratio after 1500 lives over 10 runs.}
\label{kdRatioStats}
\begin{center}
\setlength{\extrarowheight}{0.15cm}
\begin{tabular}{|c|c|c|c|}
\hline
\bf{Technique}&\bf{Average}  &\bf{Min} & \bf{Max}\\
\hline
\emph{PCWR:Yes\textunderscore PAS:Yes} &2.17:1&2.01:1&2.36:1 \\
\emph{PCWR:No\textunderscore PAS:Yes} & 2.18:1 &2.05:1&2.28:1 \\
\emph{PCWR:Yes\textunderscore PAS:No} & 0.82:1&0.78:1&0.93:1\\
\emph{PCWR:No\textunderscore PAS:No} &0.86:1&0.77:1&0.92:1\\
\hline
\end{tabular}
\end{center}
\end{table}

\begin{figure}[h]
  \begin{center}  
      \includegraphics[width=3.4in]{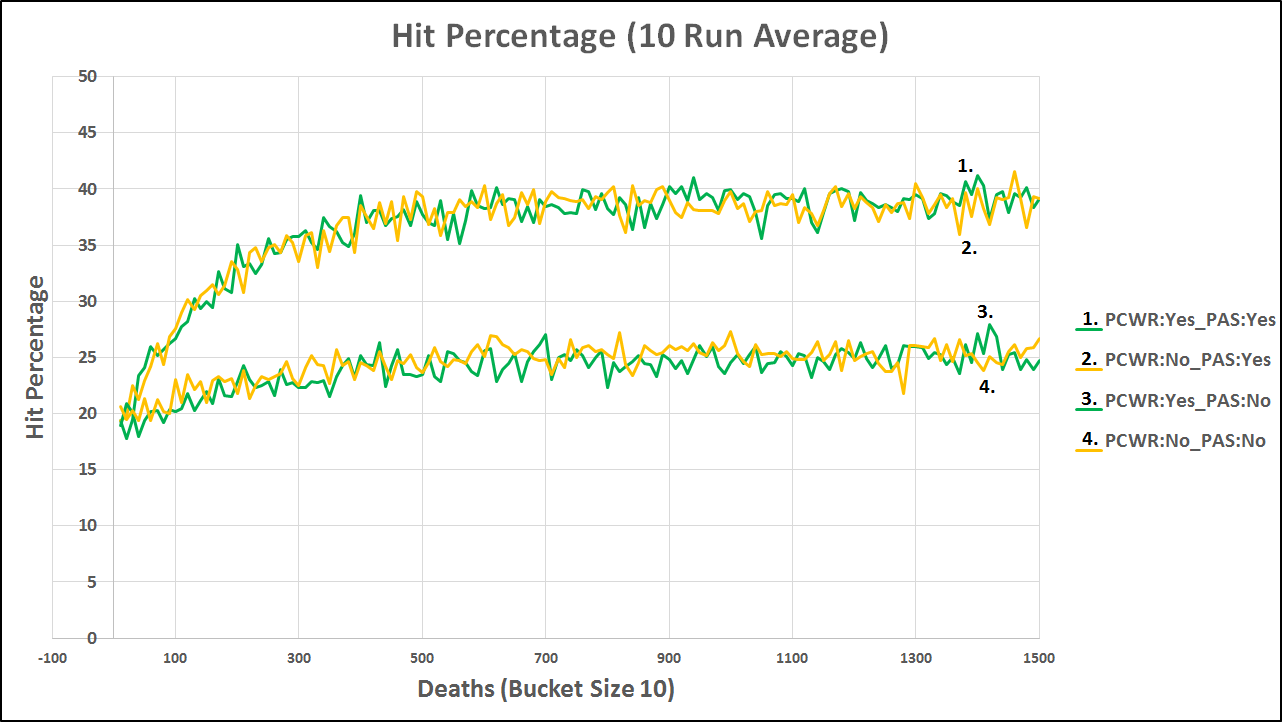}
    \caption{Percentage of hits for each variation of the architecture.}
    \label{perHits}
  \end{center}
\end{figure}

The trend for the percentage of hits over time, as learning is occurring, for each of the bots is shown in Figure \ref{perHits}. These values are averaged over the 10 games for each bot with the points on the graph also being averaged in 10-point buckets. Thus, there are 150 points on the graph to depict the hit percentages over the 1500 deaths. 

The first observation that we can make from the illustration is the separation in performance depending on whether PAS over 3 time steps is enabled or not. When actions are selected in every time step, the performance begins at about 20 percent hit accuracy and finishes just over 25 percent. However, the hit accuracy rises to 40 percent when PAS is enabled. This graph shows no clear distinction between the performance of enabling or disabling PCWR, as the averaged results fall within the same range.
\subsection{Discussion}
\begin{figure}[h]
  \begin{center}  
      \includegraphics[width=3.4in]{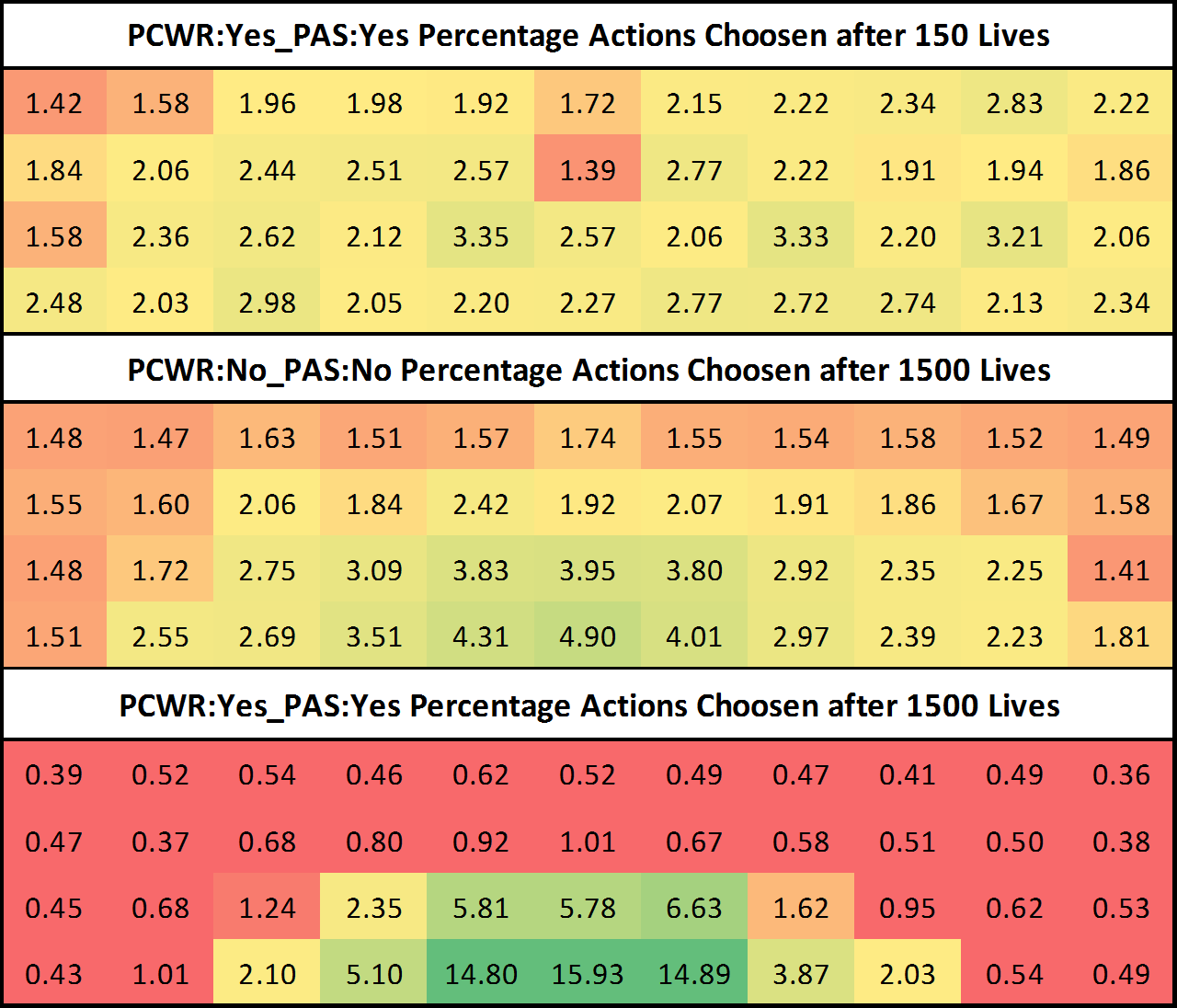}
    \caption{Percentage of overall shooting actions selected after 150 lives and after 1500 lives.}
    \label{actionPercent}
  \end{center}
\end{figure}

These results show clear evidence that the RL shooting architectures are capable of improving a bot's shooting technique over time as it learns the correct actions to take through in-game trial and error. The bot updates its state-action table, which drives its decision-making, after every shooting incident. These are carried out as mass updates once the shooting period has ended.\\
\indent The heat maps in Figure \ref{actionPercent} show the percentage of actions selected by the bot at the following stages: PCWR:Yes\textunderscore PAS:Yes after 150 lives; PCWR:No\textunderscore PAS:No after 1500 lives; PCWR:Yes\textunderscore PAS:Yes after 1500 lives. The shooting actions are those that were illustrated earlier in Figure \ref{shootingActionsNew}, which span eleven directions across the opponent at four different height levels. The heat maps clearly show the strategies that are adopted by each of the bots at the different stages of learning and the difference that selecting actions over multiple time steps can make. The diagrams show the percentage of time each shooting target was selected: at early stages (top of figure), shooting actions are widely dispersed; after learning with SARSA($\lambda$) over 1500 lives (middle graph) the bot shows a clear preference for shooting where opponents are most likely to be found and avoiding other areas such as corners; and after learning with SARSA($\lambda$) including our PAS and PCWR techniques, it shows even stronger trends in shooting preferences. 

\section{Conclusions and Future Work}
This paper has presented two techniques, Periodic Cluster-Weighted Rewarding and Persistent Action Selection, for enabling FPS bots to improve their shooting technique over time using reinforcement learning. We have demonstrated that selecting the same action over multiple time steps can lead to much better performance than when new actions are selected every time step. While our PCWR technique did achieve the highest kill streak, highest overall accuracy and highest kill-death ratio it will need further refinements to validate its usefulness as, on average, its performance was similar or worse than when the technique was disabled. On the other hand, PAS provides clear performance benefits, despite being a simple technique.\\
\indent In future work, we will refine and extend the PAS technique further to make it more broadly applicable to the problem of reward assignment in dynamic real-time environments. The state-action tables are currently stored offline after every kill or death from the bot. In the future we hope to sample from these learning stages to develop a skill balancing mechanism. \\

\bibliographystyle{IEEEtran}
\bibliography{refs}

\end{document}